\documentclass{article}

\usepackage{arxiv}

\usepackage[utf8]{inputenc} 
\usepackage[T1]{fontenc}    
\usepackage{hyperref}       
\usepackage{url}            
\usepackage{booktabs}       
\usepackage{amsfonts}       
\usepackage{nicefrac}       
\usepackage{microtype}      
\usepackage{lipsum}		
\usepackage{graphicx}
\usepackage{natbib}
\usepackage{doi}
\usepackage{amsmath}

\title{The online learning architecture with edge computing for high-level control for assisting patients}

\author{ \href{https://orcid.org/0000-0000-0000-0000}{\includegraphics[scale=0.06]{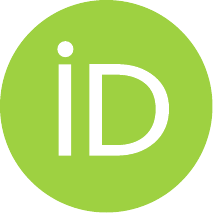}\hspace{1mm}Yue~Shi}\thanks{Yue Shi is with the School of Electronic and Electrical Engineering, University of Leeds, Leeds, UK. LS2 9JT (email:
y.shi1@leeds.ac.uk} \\
	Department of Electronic and Electrical Engineering \\
	University of Leeds\\
	Woodhouse Lane Leeds LS2 9JT \\
	\texttt{y.shi1@leeds.ac.uk} \\
	\And
	\href{https://orcid.org/0000-0000-0000-0000}{\includegraphics[scale=0.06]{orcid.pdf}\hspace{1mm} Yihui~Zhao} \\
	Bristol robotic lab, \\
	University of Bristol,\\
	Bristol, UK. \\
	\texttt{yihui.zhao@bristol.ac.uk} \\
}

\renewcommand{\shorttitle}{\textit{arXiv} Template}

\hypersetup{
pdftitle={A template for the arxiv style},
pdfsubject={q-bio.NC, q-bio.QM},
pdfauthor={Yue Shi, Yihui Zhao},
pdfkeywords={Electromyography (EMG), exoskeleton, online learning, adversarial learning, edge-computing, parallel computing},
}

\begin{document}
\maketitle

\begin{abstract}
The prevalence of mobility impairments due to conditions such as spinal cord injuries, strokes, and degenerative diseases is on the rise globally. Lower-limb exoskeletons have been increasingly recognized as a viable solution for enhancing mobility and rehabilitation for individuals with such impairments. However, existing exoskeleton control systems often suffer from limitations such as latency, lack of adaptability, and computational inefficiency. To address these challenges, this paper introduces a novel online adversarial learning architecture integrated with edge computing for high-level lower-limb exoskeleton control. In the proposed architecture, sensor data from the user is processed in real-time through edge computing nodes, which then interact with an online adversarial learning model. This model adapts to the user's specific needs and controls the exoskeleton with minimal latency. Experimental evaluations demonstrate significant improvements in control accuracy and adaptability, as well as enhanced quality-of-service (QoS) metrics. These findings indicate that the integration of online adversarial learning with edge computing offers a robust and efficient approach for the next generation of lower-limb exoskeleton control systems.
\end{abstract}


\section{Introduction}
As the global population ages, with estimates suggesting that the number of adults over 65 will triple from 524 million (~$8\%$) in 2010 to 1.5 billion (~$16\%$) in 2050 \cite{world2011strengthening}, the prevalence of age-related physical impairments such as paralysis from stroke, cerebral palsy, and spinal tumors is also expected to rise. Users of assistive technologies like exoskeletons desire devices that are intuitive to use, allowing for increased speed and assistance in daily activities. While existing exoskeletons have various applications, from lifting heavy loads to aiding in child mobility for conditions like cerebral palsy, they often lack real-time adaptability and feedback mechanisms, which hampers their effectiveness.\par

Recently, advances in human-exoskeleton cooperative control have shown promise, particularly in rehabilitation for paralyzed individuals. Electrically powered lower-limb exoskeletons have been designed for transport and handling of heavy loads. A modular architecture is employed to manage the complexity of the system design and has different operating conditions for walking, squatting, and handling loads. However, the designed machine has difficulties in maintaining equilibrium while carrying heavy loads. Exoskeletons are also used for children suffering from cerebral palsy to assist in their gait. These edge devices produce physiological gait patterns for the patients at the angle joints. Translation of gait in biped exoskeleton without using crutches is demonstrated in \cite{agrawal2017first}. A series actuator with required torque and speed for transition from sit-to-stand posture is designed in \cite{shepherd2017design}. Many exoskeletons are introduced specifically for assisting people with lower body paralysis \cite{zhang2018design}. Several other brain computer interface (BCI) technologies integrated with exoskeletons are also used for assisting the paralyzed in rehabilitation \cite{vinoj2018hybrid}. But, many of these technologies introduce additional burden to the patient and also do not provide feedback for improvement. Lack of instant control and adaptability is a major issue in most of the exoskeleton-based solutions. \par

Recently, human-exoskeleton cooperative control has gained wide acceptance in stroke rehabilitation and assistance \cite{lee2017non}. They overcome the issues with exoskeletons and provide better assistance in rehabilitation to paralyzed patients. The efficiency of these systems hugely depends on the accuracy of the captured Electromyography (EMG) signals. Recent works in full-body humanoids have highlighted their advantage over previous solutions \cite{brookes2017robots}. The current research work proposes a full-body assistive humanoid for the paralyzed by integrating the advantages offered by BCI, brain-tobrain interface and humanoid systems with software-defined network (SDN) \cite{garg2019sdn} and edge computing \cite{menon2017vehicular} technologies. SDN offers flexibility with improved network control in communication and edge computing reduces the delay in processing of the signals. These two powerful technologies provide the much-needed efficiency in the proposed system. SDN is selected with an objective of improving the network efficiency and its effective usage is implemented with universal software radio peripheral (USRP). These advantages offered by SDN are used by the proposed system to achieve simplicity, efficiency and instant control in communication. \par

A novel assistive system for paralyzed people using an SDN-powered humanoid integrated with edge computing is proposed and discussed in this article. The system consists of three modules: a human body sensor module connected to node MCU that collects electromyography (EMG), and angular motion data; an interface that utilizes edge computing and is connected to a universal software radio peripheral (USRP); and an assistive humanoid module controlled by USRP. By capturing real-time signals from human body sensors, the system allows for instantaneous control of the humanoid exoskeleton, offering a groundbreaking solution for assisting individuals with paralysis. The article proceeds to discuss the system design, theoretical analysis, hardware implementation, and suggestions for future improvements.

\section{Proposed system}
\label{sec:2}
The proposed system has three modules: 1) a human body sensor module; 2) node MCU USRP interface with edge computing module; and 3) USRP-enabled humanoid. The main function of the human body sensor module is to capture the EMG signal, angular motion, positioning information and transform it into a signal acceptable by the USRP unit. The unit has USRP motherboard and has the subsystem consisting of FPGA, DAC, ADC clock synchronization and generation. The front end is a daughterboard used for up/down signal conversion, conditioning and filtering. This flexibility allows the USRP to be used from DC to GHz application. The FPGA on USRP board performs the DSP operations allowing real analog signal to be translated into baseband digital signals. The human body sensor module uses noninvasive methods to capture the body parameter like the musicale signals from the human limb. The collected signals are then amplified using a high gain instrumentation amplifier to improve the signal strength. The signal further undergoes preprocessing and filtering. A band-pass filter is used to remove the high frequency noise. The signals are converted into frequency domain using the Walsh Hadamard transform (WHT) for feature extraction. The extracted signals are converted into digital and given to the edge-computing powered node MCU which transmits the signal via edge computing-enabled USRP. The main functionality of the node MCU with edge computing is to classify the commands and to produce actuation signal for the corresponding body part. All the decisions for control are taken by the node MCU based on the signal received via human body sensor module. \par

In the offline training phase, the users will be trained for three basic commands (sitting, standing, and sleeping). The patterns for each of these commands will be recorded to create the database. The detected sensor patterns will be mapped into these three different commands. The node MCU uses this database to make the decision regarding the action to be performed. The activation signal from the IoT-enabled edge computing is received at the receiver side by USRP trans receiver daughterboard which is connected to IoT-enabled node MCU. The node MCU decodes the received signal and passes it on to the desired part of the humanoid actuation module through the motor driver circuit. A two-level sensing mechanism is given as the feedback to the node MCU to take corrective actions. Based on the angle sensor feedback received, the node MCU makes the desired corrections on the actuation signals. The accidental fall of humanoid is detected by incorporating an accelerometer on the back of the humanoid. If the measured tilt crosses a threshold, deacceleration is provided to stabilize via edge computing-enabled USRP.\par

Now, the theoretical analysis of the data transmission in the proposed system is presented and discussed. The EMG signal processed by USRP on the humanoid can be considered as shortened at both ends of the edge forming a closed cylindrical structure. The EMG signal transmitted has an electrical component stored in the closed cylinder with power loss. The EMG signal field variation will be along three dimensions of the closed cylinder which is illustrated in Fig. 2. \par

The closed cylinder is of length b and radius c. Analysis is carried out considering the processing as lossless and then determining the quality-of-service (QoS) using the perturbation method and applying the boundary condition on the circular cylinder X = (0, a) and Y = (0, c) and EMGX = EMGY = 0 at Z = (0, b). The variation in transverse direction is represented by EMG(X, Y) and the varying amplitudes of the EMG signal as $B^{+}$ and $B^{-}$. For the EMG signal in the “x” direction and “y” direction, EMGX and EMGY are represented as
\begin{equation}
EMG(X, Y, Z) = {emg}(X, Y) \times (B^{+} e^{-j \alpha_{mn} Z} + B^{-} e^{j \alpha_{mn} Z})
\end{equation}

The permeability and permittivity of the medium are represented as $\mu$ and $\epsilon$, respectively, and the constant $ K = \omega \sqrt{\mu \epsilon}$. Now, the EMG propagation constant of $ m^{th}$ and $n^{th}$ electric components is given by

\begin{equation}
\alpha_{mn} = \sqrt{K^2 - \left( \frac{m\pi}{a} \right)^2 - \left( \frac{n\pi}{b} \right)^2}
\end{equation}

Applying the condition that $B^{+} = B^{-}$ at EMGt = 0 and at Z = 0 gives

\begin{equation}
{EMG}_t(X, Y, b) = -{EMG}_t(X, Y) B^{+} 2j \sin \alpha_{mn} d = 0
\end{equation}

The nontrivial solution $( B^+ \neq 0 )$ is the only solution which occurs for $( \alpha_{mn} d = l \pi )$ where $l = 1, 2, 3$ indicates that the EMG signal is an integer multiple. The variations of EMG at $( X, Y, Z)$ directions are represented by \( m, n, \) and \( l \) indices. Now, the EMG components for the structure is given by

\begin{equation}
O_{mnl} = \left( \frac{m\pi}{a} \right)^2 + \left( \frac{n\pi}{c} \right)^2 + \left( \frac{l\pi}{b} \right)^2
\end{equation}

The resonance frequency of USRP is given by
\begin{equation}
f_{mnl} = c \sqrt{\frac{O_{mnl}}{2 \sqrt{\mu_r \epsilon_r}}} = c \frac{\sqrt{2}}{\sqrt{\mu_r \epsilon_r}} \sqrt{\left( \frac{m\pi}{a} \right)^2 + \left( \frac{n\pi}{c} \right)^2 + \left( \frac{l\pi}{b} \right)^2}
\end{equation}

\subsection{Headings: second level}
\lipsum[5]
\begin{equation}
	\xi _{ij}(t)=P(x_{t}=i,x_{t+1}=j|y,v,w;\theta)= {\frac {\alpha _{i}(t)a^{w_t}_{ij}\beta _{j}(t+1)b^{v_{t+1}}_{j}(y_{t+1})}{\sum _{i=1}^{N} \sum _{j=1}^{N} \alpha _{i}(t)a^{w_t}_{ij}\beta _{j}(t+1)b^{v_{t+1}}_{j}(y_{t+1})}}
\end{equation}

The resonance frequency of USRP shifts depending on $(a, c)$ and $(b)$. Considering the fact that $( B^{+} = B^{-})$, the total EMG field at different directions is given by
\begin{equation}
\text{EMG}_Y = B^+ \sin\left(\frac{X}{a}\right) \left(e^{-j\alpha Z} - e^{j\alpha Z}\right)
\end{equation}

\begin{equation}
\text{EMG}_X = -B^+ \frac{Z}{\sin\left(\frac{X}{a}\right)} \left(e^{-j\alpha Z} + e^{j\alpha Z}\right)
\end{equation}

\begin{equation}
\text{EMG}_Z = j \frac{B^+}{K n a} \cos\left(\frac{X}{a}\right) \left(e^{-j\alpha Z} + e^{j\alpha Z}\right)
\end{equation}

Substituting $({EMG}_0 = -2jB^{+})$ and using $\alpha_{mn}d = l\pi$ into the above expression, we obtain
\begin{equation}
\text{EMG}_Y = \text{EMG}_0 \sin\left(\frac{X}{a}\right) \sin\left(\frac{l\pi Z}{b}\right), \label{eq:9} 
\end{equation}

\begin{equation}
\text{EMG}_X = -j\text{EMG}_0 \frac{Z}{\sin\left(\frac{X}{a}\right)} \cos\left(\frac{l\pi Z}{b}\right), \label{eq:10} 
\end{equation}

\begin{equation}
\text{EMG}_Z = j\frac{\text{EMG}_0}{K n a} \cos\left(\frac{X}{a}\right) \sin\left(\frac{l\pi Z}{b}\right). \label{eq:11}
\end{equation}

The transmitted EMG signal at USRP is given by
\begin{equation}
{USRP}_E = \frac{\epsilon}{4} \int_{V_0} {EMG}_Y {EMG}^*_Y dV = \frac{\epsilon abc}{16} {EMG}_0^2. \label{eq:12}
\end{equation}

The received EMG signal at USRP is given by
\begin{equation}
{USRP}_P = \frac{\mu}{4} \int_{V_0} (E_X E^*_X + E_Z E^*_Z) dV = \frac{\mu abc}{16} EMG_0^2 \left( \frac{1}{Z^2} + \frac{2}{K^2 n^2 a^2} \right). \label{eq:13}
\end{equation}

Substituting $ \alpha = \sqrt{K^2 - \left(\frac{\pi}{a}\right)^2}$ in $Z = \frac{Kn}{\alpha}$ and then substituting the result in \eqref{eq:13}, we obtain
\begin{equation}
{USRP}_P = \frac{1}{Z^2} + \frac{2}{K^2 n^2 a^2} = \frac{\alpha^2 + \left(\frac{\pi}{a}\right)^2}{K^2 n^2} = \frac{1}{n^2} = \frac{\epsilon}{\mu}. \label{eq:14}
\end{equation}

Considering $R_S$ as the resistivity of USRP and ${EMG}_T$ as the tangential electric field at the closed cylindrical structure, the power lost in the transmission is given by
\begin{equation}
P_{LT} = \frac{R_S}{2} \int | {EMG}_t |^2 dS. \label{eq:15}
\end{equation}

Using the value of $({EMG}_X)$ and $({EMG}_Y)$ in the above equation, we obtain
\begin{align}
P_{LT} &= \frac{R_S}{2} \left[ 2 \int_{Y=0}^{c} \int_{X=0}^{a} | \text{EMG}_X(Z = 0) |^2 + 2 \int_{Z=0}^{b} \int_{Y=0}^{c} | \text{EMG}_Z(X = 0) |^2 \right. \nonumber \\
&\left. + 2 \int_{Z=0}^{b} \int_{X=0}^{a} | \text{EMG}_X(Y = 0) |^2 + | \text{EMG}_Z(Y = 0) |^2 dY dZ \right] \nonumber \\
&= \frac{R_S \text{EMG}_0^2 \lambda^2}{8 n^2} \left[ \frac{l^2 a c}{b^2} + \frac{c b}{a^2} + \frac{l^2 a^2}{b} + \frac{b^2}{a} \right]. \label{eq:16}
\end{align}

Here, the symmetry of the circular cylinder structure at$ ( X = 0 )$, $( Y = 0 )$, and $( Z = 0 )$ is accumulated till $( X = a )$, $( Y = c )$, and $( Z = b )$.

The QoS at USRP is obtained as
\begin{equation}
Q_{USRP}(TX) = \frac{2 \omega_0 S D R E}{P_L} = \frac{K^3 a b c n^4}{8 R_S} \left[ \frac{2 l^2 a^3 c}{b^2} + \frac{2 c b^3}{a^2} + \frac{2 l^2 a^3 b}{b^2} + \frac{2 a b^3}{a^2} \right]. \label{eq:17}
\end{equation}

The effective transmission conductivity of USRP is given by
\begin{equation}
\sigma = \omega \epsilon = \omega \epsilon_r \epsilon_0 \tan \delta, \quad \epsilon = \epsilon_r \epsilon_0 (1 - j \tan \delta). \label{eq:18}
\end{equation}

where \( \tan \delta \) denotes loss tangent.

The total received power at the received side is given by
\begin{equation}
P_R = \frac{1}{2} \int_{V_0} {EMG} \cdot {EMG}^* dV = \frac{\omega \epsilon''}{2} \int_{V_0} | {EMG} |^2 dV = \frac{a c b \omega \epsilon'' | {EMG} |^2}{8},
\end{equation}

and
\begin{equation}
Q_{{USRP}}(RX) = \frac{2 \omega {USRP}_E}{P_R} = \frac{1}{\tan \delta}. \label{eq:19}
\end{equation}

When both USRP transmitted and received signal loss exist, the total power loss is \( (P_T + P_R) \). The total QoS of USRP is
\begin{equation}
Q_{{USRP}} = \left( \frac{1}{Q_{{USRP}}(TX)} + \frac{1}{Q_{{USRP}}(RX)} \right)^{-1}. \label{eq:20}
\end{equation}

Fig. 3 presents the architecture of the proposed system. Here, the transmitter side mainly includes the sensor data generation module, the cyclic-prefix orthogonal frequencydivision multiplexing (CP-OFDMA) module and the raised cosine transmitter filter. The sensor data module generates the bits for each frame and the CP-OFDMA module modulates the bits into CP-OFDMA symbols. Raised cosine transmit filter (RCTF) uses a rolloff factor at 0.5 to up sample the CP-OFDMA symbol by two. The channel used is additive white Gaussian noise (AWGM) with frequency offset and variable timing drift. At the receiver side the input data is received from the two USRP devices. The data captured has a base band signal with sampling frequency of 200 KHz. With automatic gate control (AGC), the derived amplitude level is adjusted for the received signal. The cosine receive filter with root raised, decimates the received signal by two and set a rolloff factor of 0.5. Coarse frequency compensation (CFC) calculates an approximate offset frequency of the input signal and rectifies it. Fine frequency compensation (FFC) compensates the phase and the residual frequency offset. With timing recovery, the recovered stroke timing is used to resample the received signal such that the correct symbol decision is made for the optimized sampling frequency. Decoding module demodulates the received signal and aligns the frame bit boundary. It also resolves the ambiguity of the carrier phase caused by the FFC sub system. The three Boolean signals, bit 1 and bit 2 and the valid data are the output from the receiver. To speed up the computation, frame-based signal processing is used. In downstream bit processing, the sample-based signal is converted to frame-based signal using bit message decoding. The output demodulated bit 1 and bit 2 are valid only when the valid data is high. The bit message decoding block uses valid data to fill the delay string of bit 1 and bit 2. \par

The first block AGC is used to ensure a fixed and stable input to the timing and frequency recovery block. This is done by setting the amplitude of the input CFC systems as 1/up sampling factor and to make sure that the phase and timing error detection gain remain constant with respect to time. The positioning of AGC before the cosine receiver is to make sure that the over sampling factor of four is used to measure the amplitude of the signal. This ensures improvement in the accuracy of the estimation. The RCTF helps with optimized SNR for the transmitted waveform using matched filtering and assures smooth downstream signal processing. The CFC corrects the required signal with approximate estimation of the frequency offset. The phase and frequency offset of the base band signal is then calculated. The received signal is raised to the power of four using the cascaded product block. The tones at the four-timing multiplication factor of the frequency offset is estimated. This estimated factor is divided by four, and the frequency offset is corrected in the original received signal. The residual frequency offset is then removed by FFC which is implemented using a phase locked loop (PLL). A correlation-based technique is used for estimating the frequency compensation with FFC which saves the hardware resources. The circuit speed is ensured by implementing the pipeline register. The phase offset and the residual frequency offset are compensated using PLL in the input signal. The maximum likelihood phase error detection (PED) will generate the error phase shift. The integral loop filter with automatic timing removes the error signal and feeds it to the phase estimation block. The phase estimation block generates the exponential signal to correct the phase offset and the residual frequency at the output signal of CFC. The timing error in the received signal is corrected by using the PLL. On average, one output sample is generated for two given input samples. It also gives a timing strobe valid data signal that has input sample rate as reference. Under normal operation, the value of the strobe signal is alternate zeroes and ones in sequence. The data decoding module performs the carrier phase error resolution and frame synchronization. Here, the frequency synchronization is achieved by matched filtering. The AWGM channel is enabled with variable delay and frequency offset. It applies frequency offset and a preset phase offset to the signal which has to be transmitted and then adds the variable delay. The next section presents the discussion on the results achieved with the proposed method.

\section{RESULTS AND DISCUSSION}
The hardware implementation details and the results achieved with the proposed method are presented in this sec- tion. The human body sensor module consists of an EMG sensors to capture EMG, positioning and angular motion information. The collected signals are amplified by using high gain instrumentation amplifier to improve the signal strength. The signal further undergoes preprocessing and filtering. A band-pass filter is used for removing high frequency noise. The signals are then converted into frequency domain using WHT for feature extraction. The extracted signals are converted into digital and given to the edge-computing powered node MCU which transmits the signal via edge computing-enabled USRP. The main functionality of the node MCU with IoT-enabled edge computing is to classify the commands and to produce actuation signals for the corresponding body part. All the decisions regarding control are taken by the node MCU based on the signal received via human body sensor module. An exoskeleton integrated with humanoid hand is used for the experiment and to validate the proposed system.\par

\begin{figure}[]   
    \centering  
    \includegraphics[width=\textwidth]{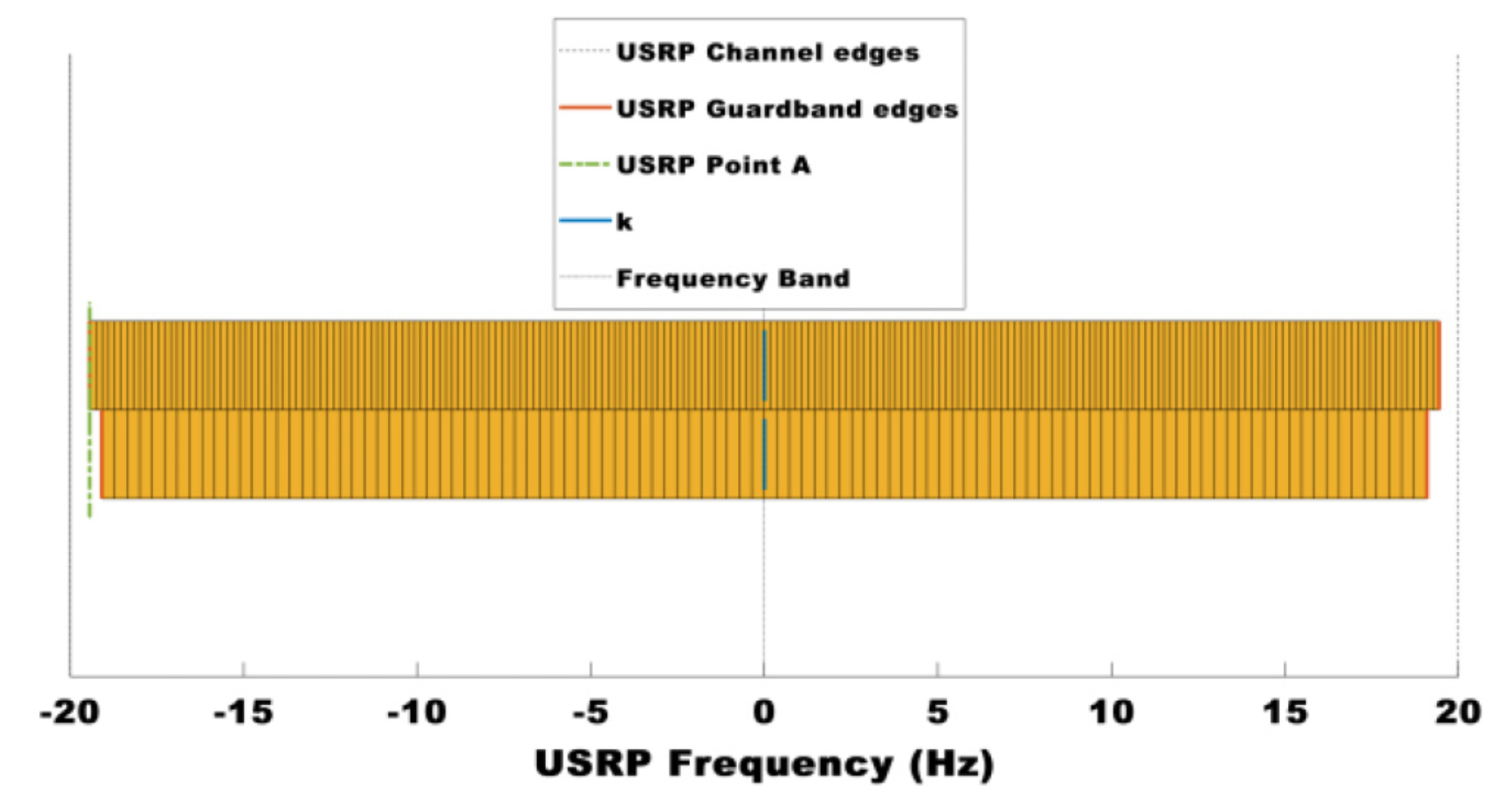}   
    \caption{ Reference components of USRP.}  
    \label{fig:1}  
\end{figure}

\begin{figure}[]   
    \centering  
    \includegraphics[width=\textwidth]{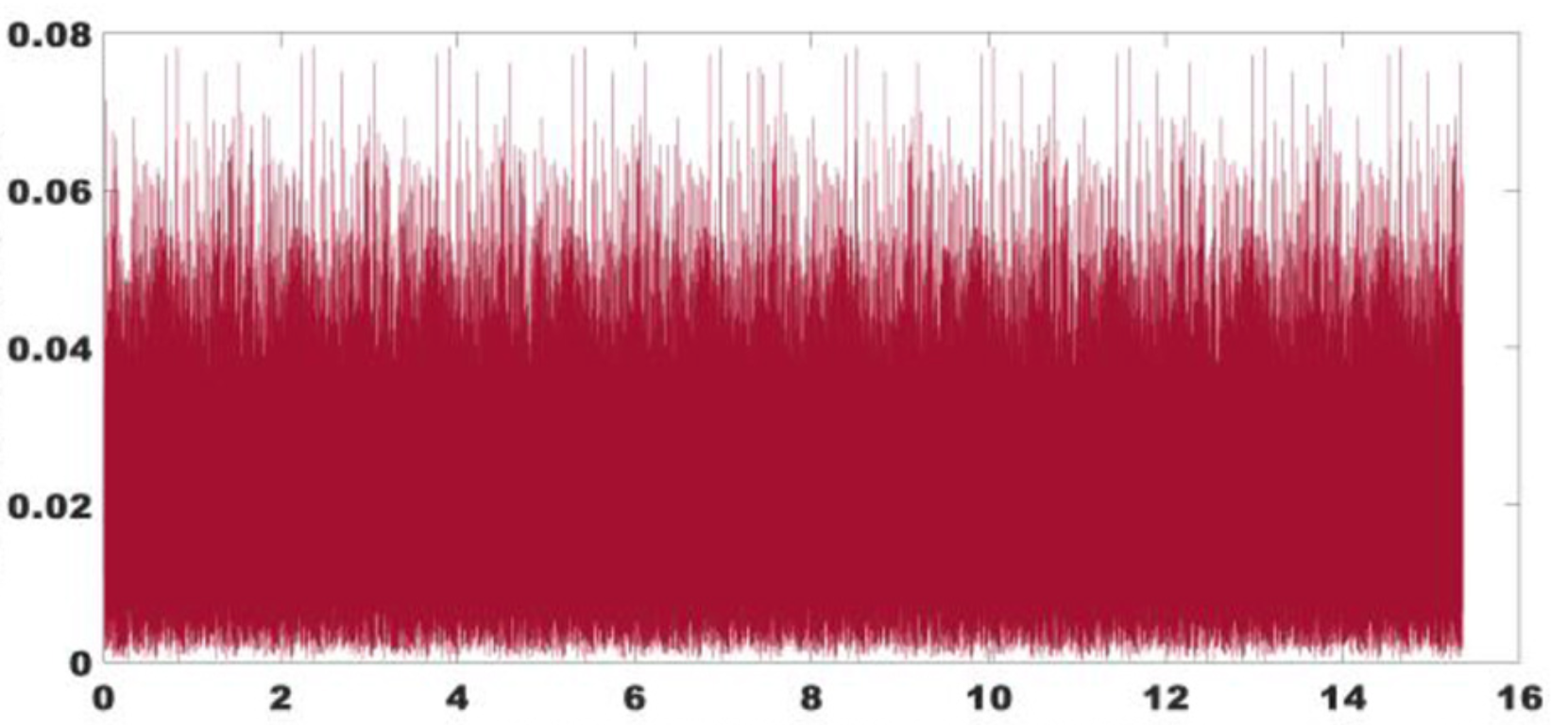}   
    \caption{Amplitude variations with respect to the frequency of EEG electrical
components}  
    \label{fig:2}  
\end{figure}

Fig. \ref{fig:1} presents the various reference components of the USRP used with the proposed system. The frequency range used in the transmission is between $-20$ and 20 Hz. This is presented to depict the fundamental setting in the data trans- mission with USRP. In the proposed system, two USRPs are used at the transmitter side and two at the receiver side. They provide simple and efficient connectivity between the sensor module and the assistive humanoid. Further, the USRP enabled proposed system achieves very good QoS in different humanoid hand postures. The amplitude variations with respect to frequency of EMG electrical component is shown in Fig. \ref{fig:2}. In the proposed system, with the humanoid hand, two movements, hand roll up and hand roll down are performed. Amplitude variations with respect to the frequency of EMG electrical components will help in the estimation of the correct decision regarding the hand movement by the humanoid. The correct hand posture is selected based on the matching of the signal parameters with the already recorded values. The nearest matching value is selected and the corresponding hand movement is performed. \par

\begin{figure}[]   
    \centering  
    \includegraphics[width=\textwidth]{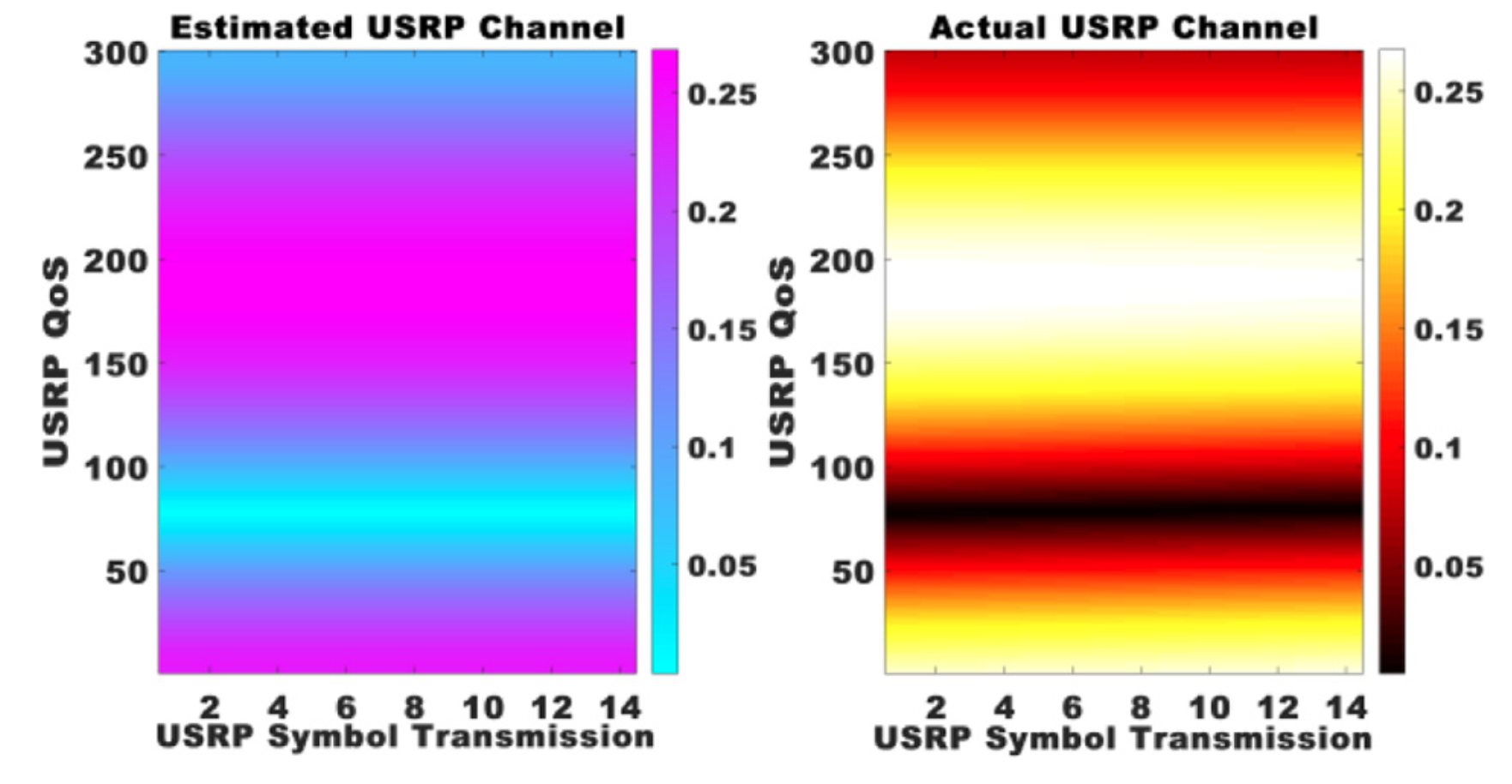}   
    \caption{Estimated and actual USRP channel estimation.}  
    \label{fig:3}  
\end{figure}

Fig. \ref{fig:3} gives the actual and estimated USRP channel for transmission and reception. This result is also very important for making the correct decision on the appropriate hand movement at the receiver end. From the results obtained with the proposed system, it is observed that the actual channel converges with the estimated channel which helps to make the right decision on the selection of the exact hand movement by the system.

\begin{figure}[]   
    \centering  
    \includegraphics[width=\textwidth]{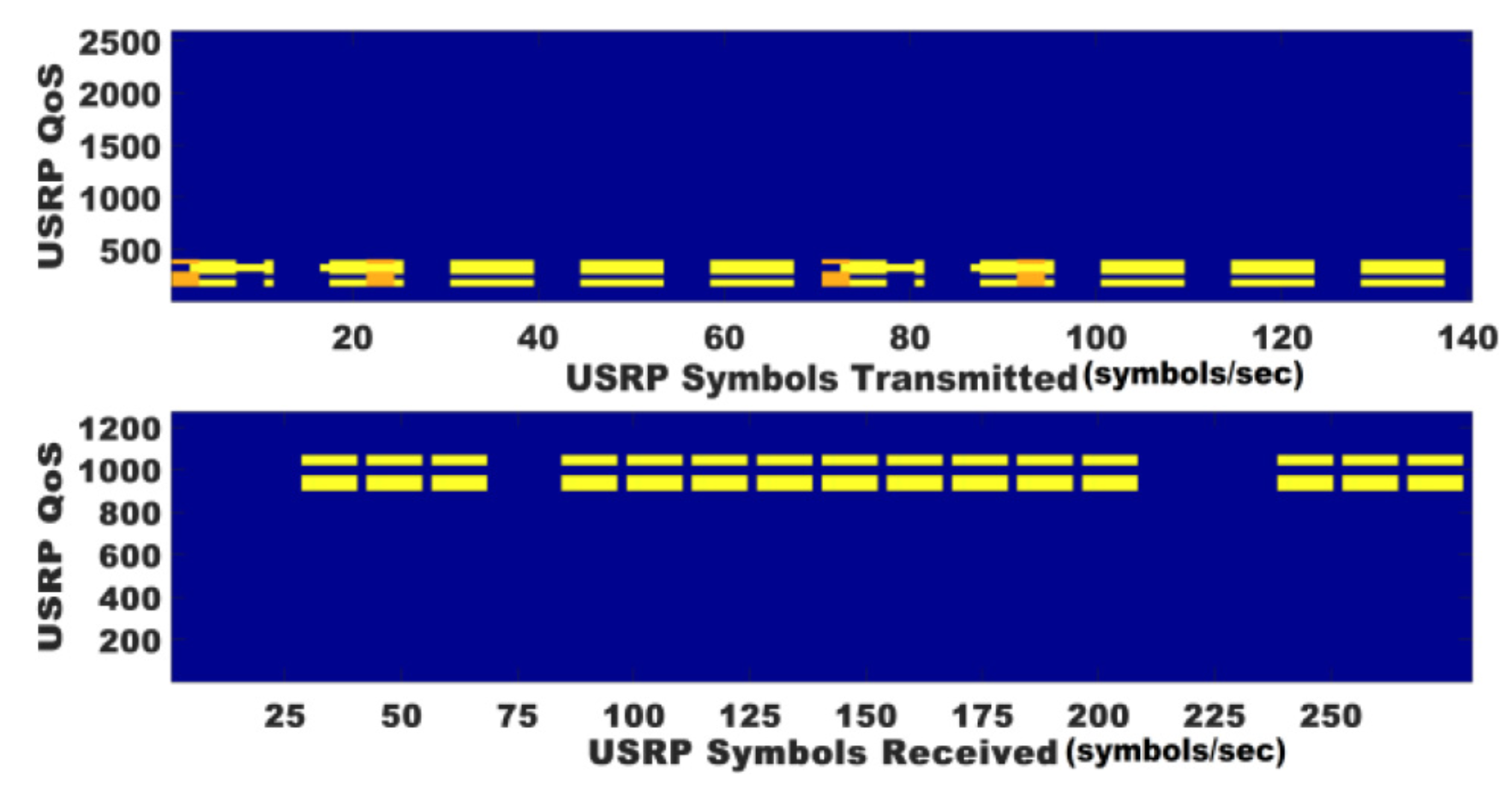}   
    \caption{USRP QoS variations with respect to symbols transmitted/received.}  
    \label{fig:4}  
\end{figure}

\begin{figure}[]   
    \centering  
    \includegraphics[width=\textwidth]{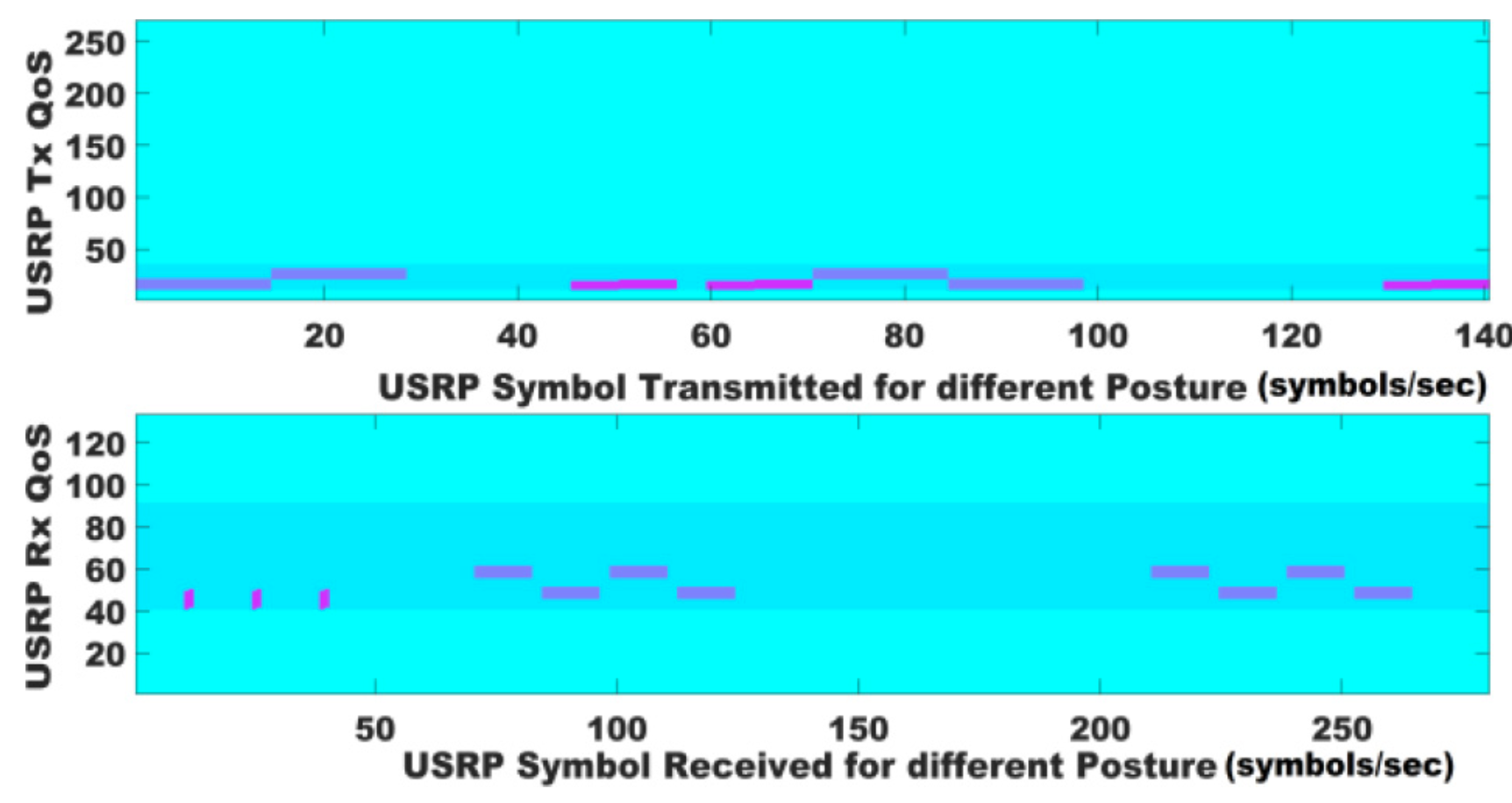}   
    \caption{USRP transmission/reception QoS with respect to symbols transmitted/received for different postures.}  
    \label{fig:5}  
\end{figure}

\begin{figure}[]   
    \centering  
    \includegraphics[width=\textwidth]{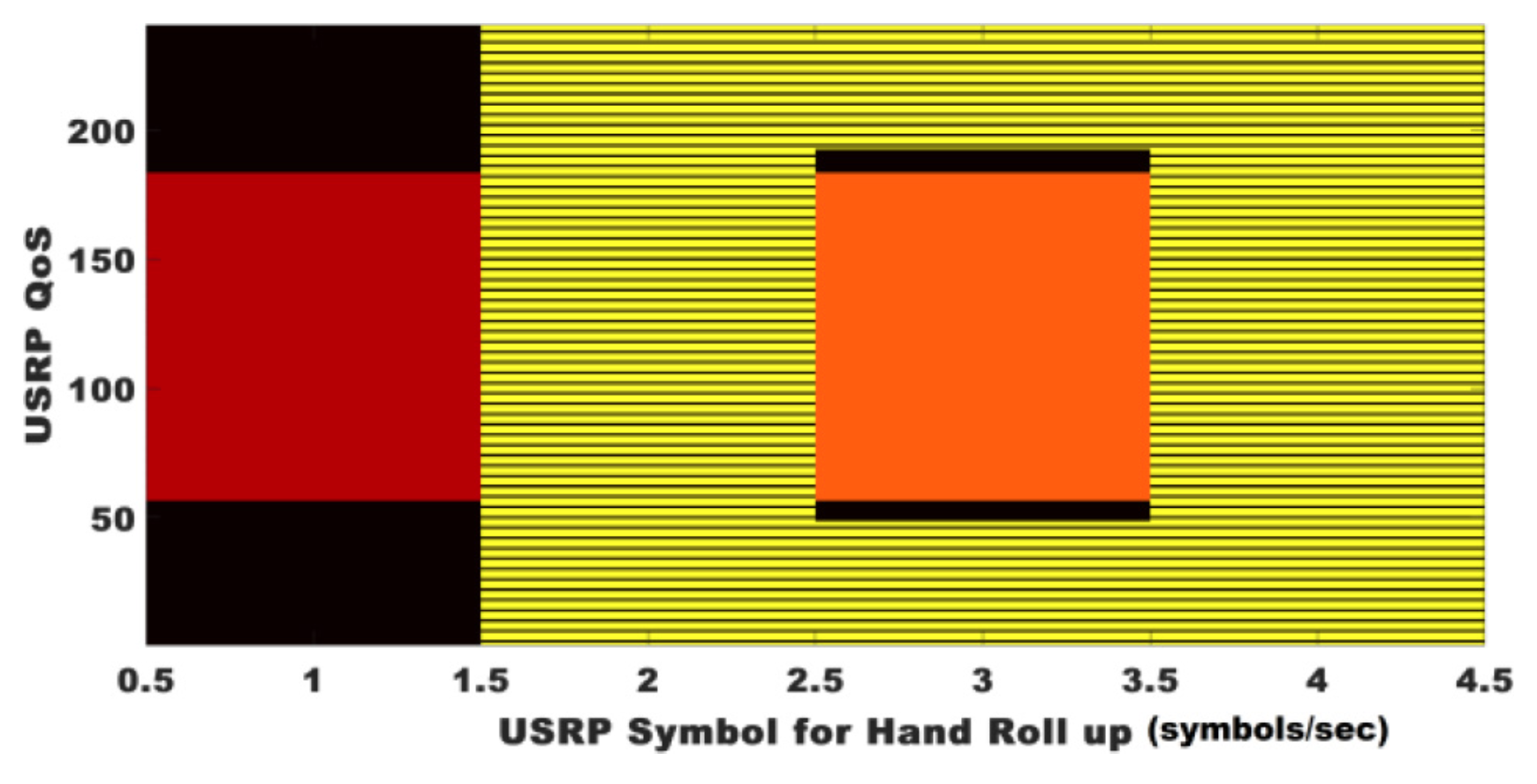}   
    \caption{USRP QoS with hand roll up.}  
    \label{fig:6}  
\end{figure}

Fig. \ref{fig:4} presents the USRP QoS variations with respect to symbols transmitted and received. The QoS of the signal at the transmitter is shown in the top half of the figure and the QoS at the receiver is shown at the bottom half. It is quite evident from the results plotted in the graph that the proposed system with USRP achieves good performance at the humanoid end. The system offers very good performance in communication and thus contributes to the overall success in the working of the assistive humanoid system. The USRP QoS variations with respect to symbols transmitted and received for different hand movements, hand roll-up and hand roll down is shown in Fig. \ref{fig:5}. From the graph, it is evident that a higher QoS is achieved at the receiving end for different hand movements, which justifies the excellent performance of the proposed system. Fig. \ref{fig:6} presents the QoS of the proposed system with the hand roll up movement and Fig. \ref{fig:7} presents the QoS with the hand roll down movement. It is evident from the graph that the actual QoS achieved by proposed system while controlling the humanoid in various movements is almost equivalent to the estimated QoS and this confirms the high performance achieved by the proposed SDN powered humanoid assistive system.

\begin{figure}[]   
    \centering  
    \includegraphics[width=\textwidth]{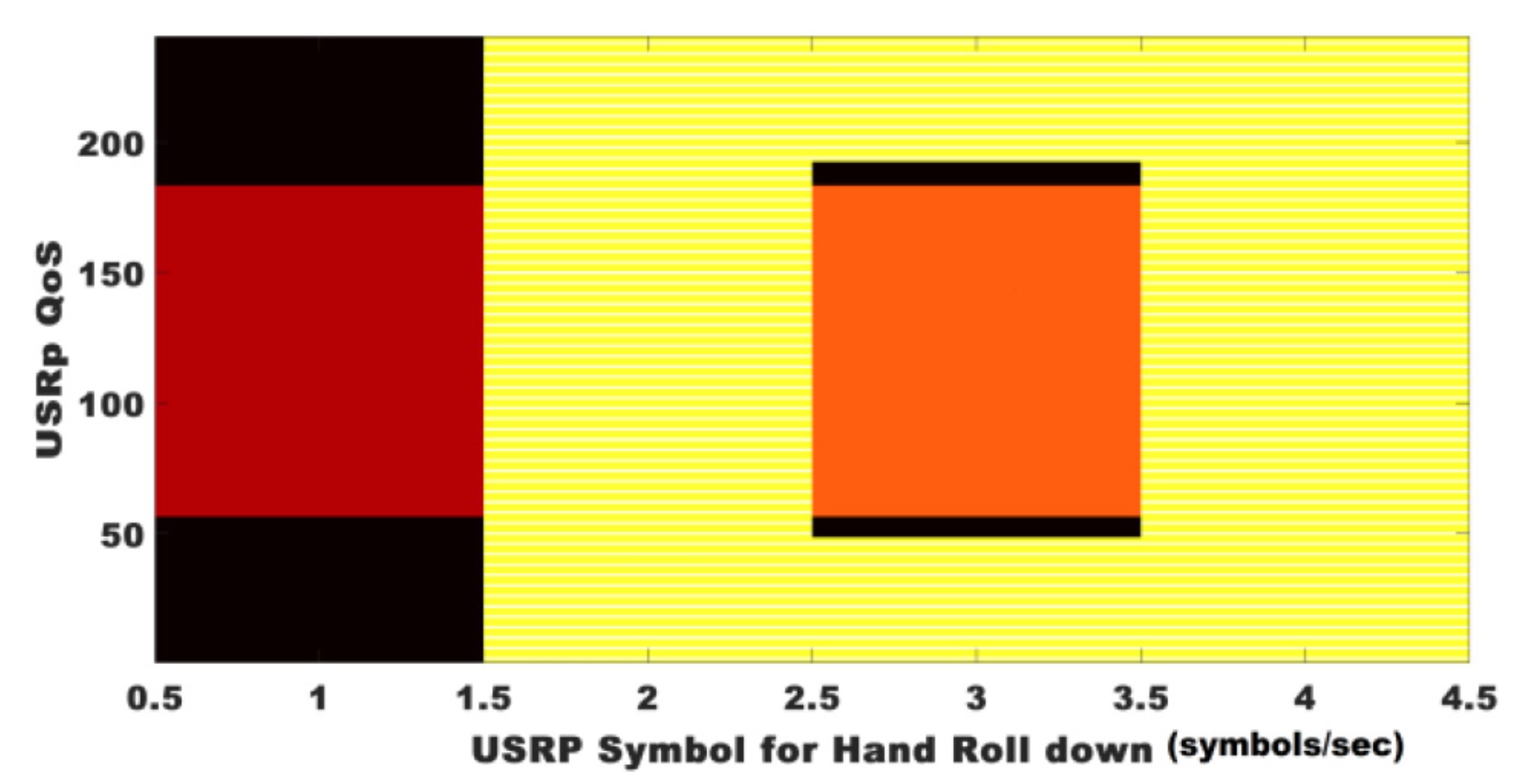}   
    \caption{Estimated and actual USRP channel estimation.}  
    \label{fig:7}  
\end{figure}

Fig. 12 presents the variations in EEG components obtained for different postures. The graph is plotted based on sample results obtained with the proposed system. From the graph, it is observed that the hand roll-up posture is selected in the current scenario for activating the humanoid based on the peak component value observed in the result. The proposed system makes the right posture decisions accurately based on the peak component values achieved. Thus, the proposed SDN-enabled assistive humanoid system is efficient and provides instant control and assistance in rehabilitation of the paralyzed.

\begin{figure}[]   
    \centering  
    \includegraphics[width=\textwidth]{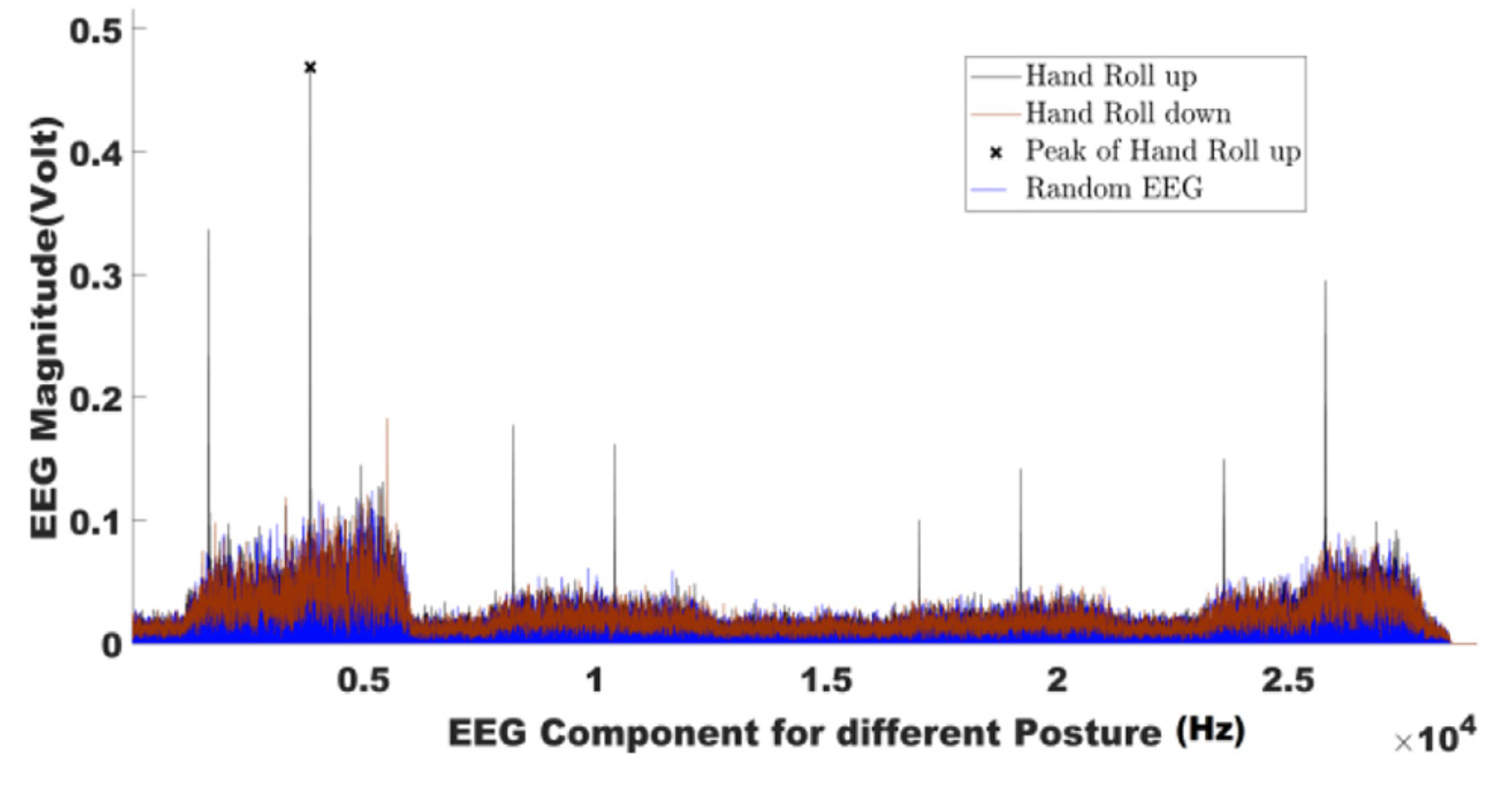}   
    \caption{EMG magnitude variation in the receiver system.}  
    \label{fig:7}  
\end{figure}

\section{Conclusion}
This article proposed a novel assistive system for paralyzed people using an SDN-USRP powered humanoid integrated with edge computing. The system consists of three modules, a human body sensor module connected to node MCU to collect EMG and angular motion data, a node MCU-USRP interface with edge computing and USRP- enabled humanoid. The signals captured by the human body sensors in turn helped to control the humanoid and this system is used for assisting the paralyzed. The proposed system overcomes the limitations of normal exoskeletons used for paralysis assistance. The experimental setup was done for controlling a humanoid hand and results showed high QoS for hand roll-up and roll-down posture. The results showed that the SDN-enabled body sensor module is an efficient method for providing instant control for the assistance and rehabilitation of the paralyzed. In the future, the analysis of the system performance needs to be done with the full body humanoid and also more efficient deep learning algorithms can be integrated to improve the accuracy of movement prediction and decision making in the humanoid.

\bibliographystyle{unsrtnat}
\bibliography{references}  






\end{document}